\documentclass[sigconf]{acmart}

\usepackage{booktabs} 
\usepackage{subcaption}
\usepackage{graphicx}
\usepackage{multirow}

\setcopyright{rightsretained}

\acmConference[]{}

\begin{document}
\title{Utility in Fashion with implicit feedback}

\author{Vikram Garg}
\affiliation{%
  \institution{Myntra Designs Pvt. Ltd.}
 }
\email{vikram.garg@myntra.com}

\author{Girish Sathyanarayana}
\affiliation{%
  \institution{Myntra Designs Pvt. Ltd.}
 }
\email{girish.sathyanarayana@myntra.com}

\author{Sumit Borar}
\affiliation{%
  \institution{Myntra Designs Pvt. Ltd.}
 }
\email{sumit.borar@myntra.com}

\author{Aruna Rajan}
\affiliation{%
  \institution{Myntra Designs Pvt. Ltd.}
 }
\email{aruna.rajan@myntra.com}

\begin{abstract}
Fashion preference is a fuzzy concept that depends on customer taste, prevailing norms in fashion product/style, henceforth used interchangeably, and a customer's perception of utility or fashionability, yet fashion e-retail relies on algorithmically generated search and recommendation systems that process structured data and images to best match customer preference. Retailers study tastes solely as a function of what sold vs what did not, and take it to represent customer preference. Such explicit modeling, however, belies the underlying user preference, which is a complicated interplay of preference and commercials such as brand, price point, promotions, other sale events, and competitor push/marketing. It is hard to infer a notion of utility or even customer preference by looking at sales data.  

In search and recommendation systems for fashion e-retail, customer preference is implicitly derived by user-user similarity or item-item similarity. In this work, we aim to derive a metric that separates the buying preferences of users from the commercials of the merchandise (price, promotions, etc). We extend our earlier work on explicit signals to gauge sellability or preference \cite{garg2016sales} with implicit signals from user behaviour.

\end{abstract}

%
%


\maketitle

\section{Introduction}

On an e-commerce platform, user's purchase decision is heavily influenced by price, discounts, brand, product attributes and visual representation of the product. Merchandising bias, hence, is a huge driver for user behaviour, especially in e-retail in India where sales are primarily discount led. It's been observed \cite{isabella2012influence} that product-sells get exponentially impacted with rise in discounts. Similarly daily rate of sales of products is inversely related to price bands of the competing product. Further users cluster many brands together and see them as similar and hence these brands behave as substitute goods for users. \cite{isabella2012influence} These merchandising parameters pose challenges in understanding the true value (to a customer) in a fashion product, regardless of its commercials. 

In this paper, we propose an efficient way to model user sessions to learn this value, or inherent customer preference for fashion products, free of merchandising bias. We use click-stream logs from Myntra - the biggest fashion e-retailer in India to calculate implicit user preferences. The rest of paper is organized as follows. Section 2 lists related works and prior art. In Section 3 we present methods of capturing the user behaviour and we describe the data preparations. In Section 4 we presents the experiments around the proposed approaches which showcase the effectiveness. In Section 5 we show the conclusion on the work and mention the future steps. 

\section{Related Work}

Effective estimation of aesthetics preference of images has several applications in e-retail industry varying from what to display on the search and recommendation to assortment decisions to get the relevant products in road shows. The majority of research in preference estimation is based on labelled data \cite{murray2012ava, lu2015deep} and using image embeddings generated by popular deep learning models \cite{russakovsky2015imagenet,he2016deep,szegedy2015going,russakovsky2015imagenet}. Research has been done by indirectly inferencing the aesthetics preference of the images from the various other labels from standard datasets such as AVA, CUHK, CUHKPQ, MIRFLICKR \cite{lu2014rapid, murray2012ava, ke2006design, luo2011content, muller2010experimental}. Few of them also focus on handcrafted cues such as color space \cite{nishiyama2011aesthetic, o2011color}, image texture \cite{datta2006studying, ke2006design}, content information \cite{luo2011content,dhar2011high} and few have worked using generic image features such as SIFT and Fisher Vector \cite{marchesotti2013learning, marchesotti2011assessing}. \cite{zhang2015daily} has done work on time aware recommender system using items affinity. \cite{he2016ups} models a mixture of time aware and visual aware recommender.  All the above methods relies on the labelled data (explicit feedback) or transaction data (implicit feedback) to model the visual aesthetics. labelled data (explicit feedback) is subjective and time dependent. The transaction data  (implicit feedback) in an e-retail/commerce setting is generally biased because customer transactions are not only influenced by visual aesthetics of the product but also (to a large extent) due to various merchandising bias such as discounts, brand value, inorganic push on list views etc. \cite{garg2016sales} learns normalized sales numbers using image similarity constraint. Present work look at the problem from a different point of view by modelling the implicit feedback from user behaviour rather than transaction data. 

\section{Modeling User Behavior}

Typically an ecommerce platform monitors and logs all user events such as user searches, clicks, add to cart, filters, orders etc along with the timestamps of such activities. Aggregation of such user logs are called user session. For the purpose of this work a user session in our platform begins when a user arrives on the website or mobile application and ends when we observe a 30 minutes of inactivity. We gather this data from our platform, Myntra, for about 30 million users every month. 

\subsection{Click based method}
\begin{figure}
  \includegraphics[width=\linewidth]{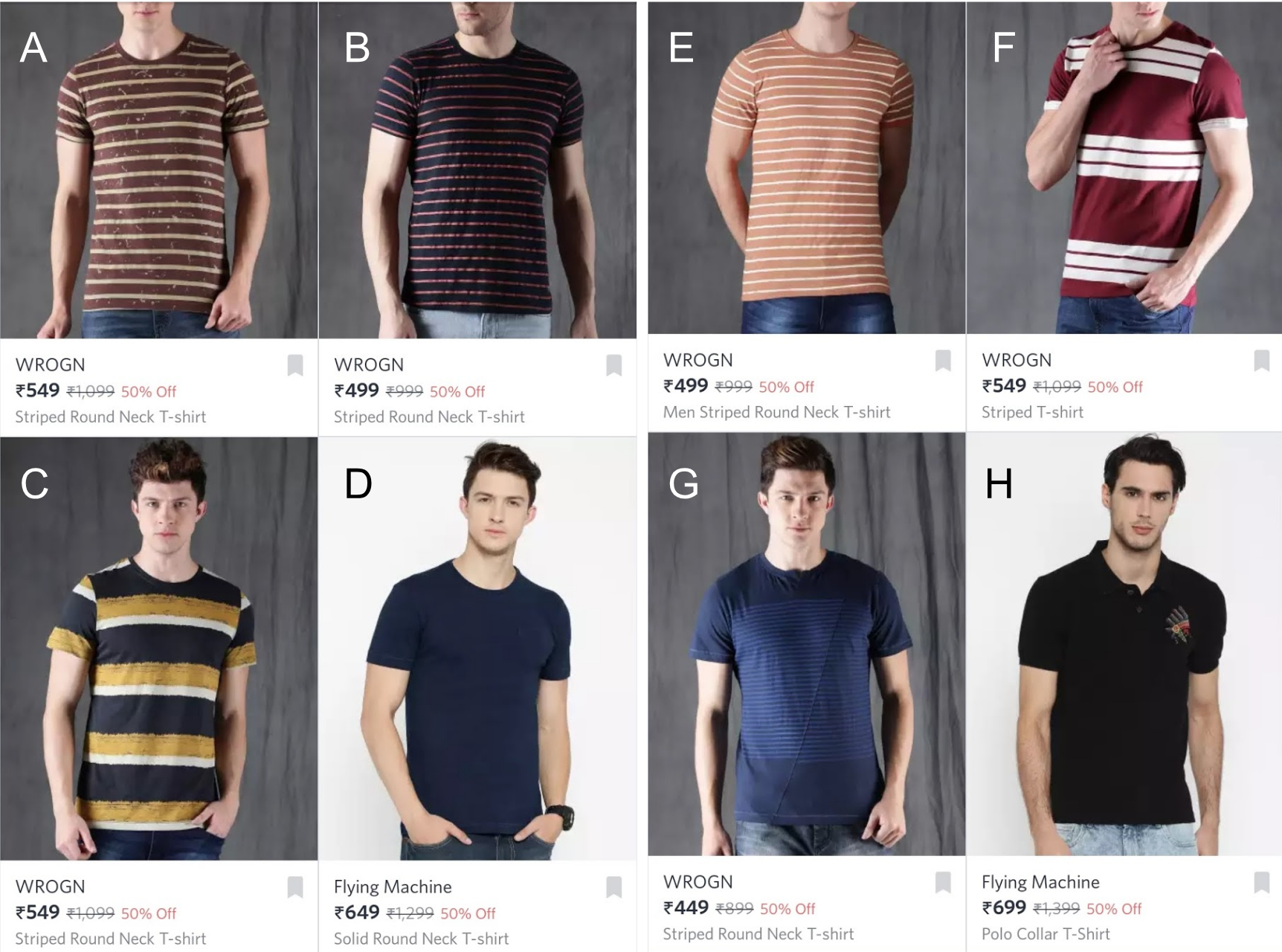}
  \caption{Fashion products displayed on our mobile application in the given order (A to H i.e. 4x4 products per panel)}
  \label{fig:tshirts}
\end{figure}

Figure \ref{fig:tshirts} shows a search session where user searched for particular type of product. Let's see an example where he/she is looking for round neck men t-shirts. Let's say in this particular search session users ends up clicking product G. Since our mobile application show 4x4 products per panel on screen, this means that user has skipped products from A to F. In other words given that all the products in this search session belong to very similar merchandising parameters user shows a preference towards product G over products from A to F. Since merchandising parameters remains similar the aesthetics preference of product G is the only factor that guides a user to click on it.  We categorized all such preferred products as s1 and all skipped products as s2. Products shown after the selected product G are not taken into consideration (products H and onwards in this example). Thus this user-session generates a list (L) containing tuples/pair (s1, s2): [(G, A), (G, B), (G, C) \dots (G, F)]. Before finalizing these tuples/pair, we explicitly apply an additional layer of filtering that ensures that all the products belong to very similar brands, similar average selling prices, and similar discounts. This essentially ensures the implicit normalization for the merchandising. E.g. product D was selling at about 44\% higher price than that of product G. Our additional layer of filter remove (G, D) pair form list L.

\subsection{Data preparation}
We scanned through the user session as described above for a given time range and for men tshirts. We get approximately 4 millions such (s1, s2) pairs from all our unsigned or signed in users. We sum the number of times  (s1, s2) occurred together and get a list of style pairs in the $(s1^{(i)}, s2^{(i)} , Counts^{(i)} )$ format where \newline

$Counts^{(i)} \hspace{0.1cm} = \hspace{0.1cm}\Sigma_{i=1}^N \hspace{0.1cm} (1 \hspace{0.1cm} if \hspace{0.1cm} s1^{(i)} \hspace{0.1cm} is \hspace{0.1cm} clicked \hspace{0.1cm} and \hspace{0.1cm} s2^{(i)} \hspace{0.1cm} is \hspace{0.1cm}skipped \newline 
and \hspace{0.1cm} -1 \hspace{0.1cm} if \hspace{0.1cm} s2^{(i)} \hspace{0.1cm}  is \hspace{0.1cm} clicked \hspace{0.1cm} and \hspace{0.1cm} s1^{(i)} \hspace{0.1cm} is \hspace{0.1cm} skipped)$\newline
for all $(s1^{(i)} , s2^{(i)})$ where i is limited by number of user sessions $S_i$, i = 1...N. \newline 

if $Counts^{(i)}$ is negative then it signifies that $s2^{(i)}$ was preferred by larger number of users rather than $s1^{(i)}$ and we simply switch the tuple values as $(s2^{(i)}, s1^{(i)} , |Counts^{(i)}| )$

About half of the pairs occurred only once ($Counts^{(i)}$ is 1) and  about 98\% of them occurred less than 40 times ($Counts^{(i)}$ < 40). This leads us to about 200k tuples $(s1^{(i)} , s2^{(i)})$ with counts more than 40 containing about 11k unique styles. We also got the styles independent transaction numbers. 

We compared these (s1, s2) tuples in terms of their independent transaction on our platform. We found that if the product s1 was preferred over s2 in terms of clicks in a given search session then that product approximately 70\% of times $s1^{(i)}$ was more sold on than $s2^{(i)}$. We observed similar correlations for \textit{AddToCart} of these products. \textit{AddToCart} is a strong signal from user because it signifies that users are interested in these products but have not purchased them so far.  It's important to note that $(s1^{(i)} , s2^{(i)})$ tuple gets generated if they both occur in same search session. But $s1^{(i)}$ and $s2^{(i)}$\'s purchase number are independent of the fact that they appear in same user session or not. 

\begin{figure}[!h]
    \centering
  \begin{subfigure}[b]{0.9\linewidth}
    \includegraphics[width=\linewidth]{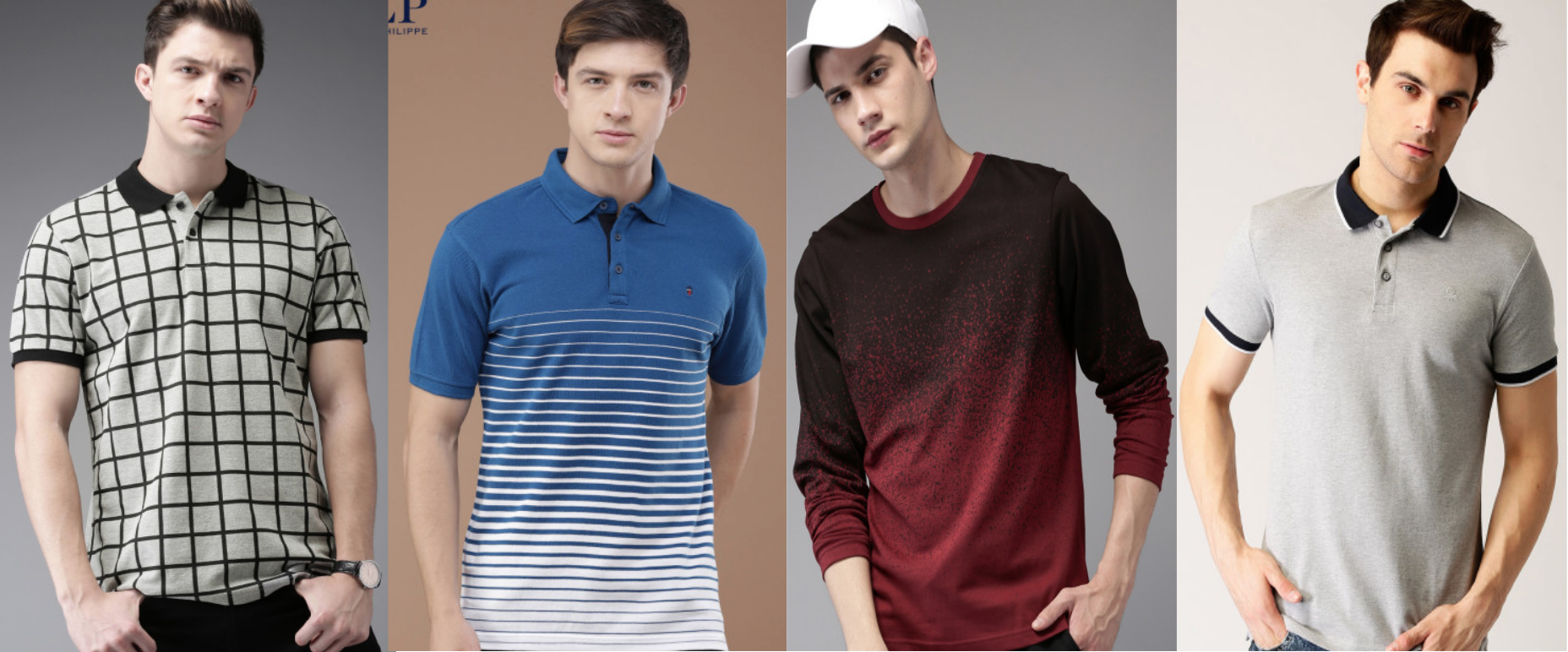}
    \caption{Preferred Styles}
  \end{subfigure}
    \begin{subfigure}[b]{0.9\linewidth}
    \includegraphics[width=\linewidth]{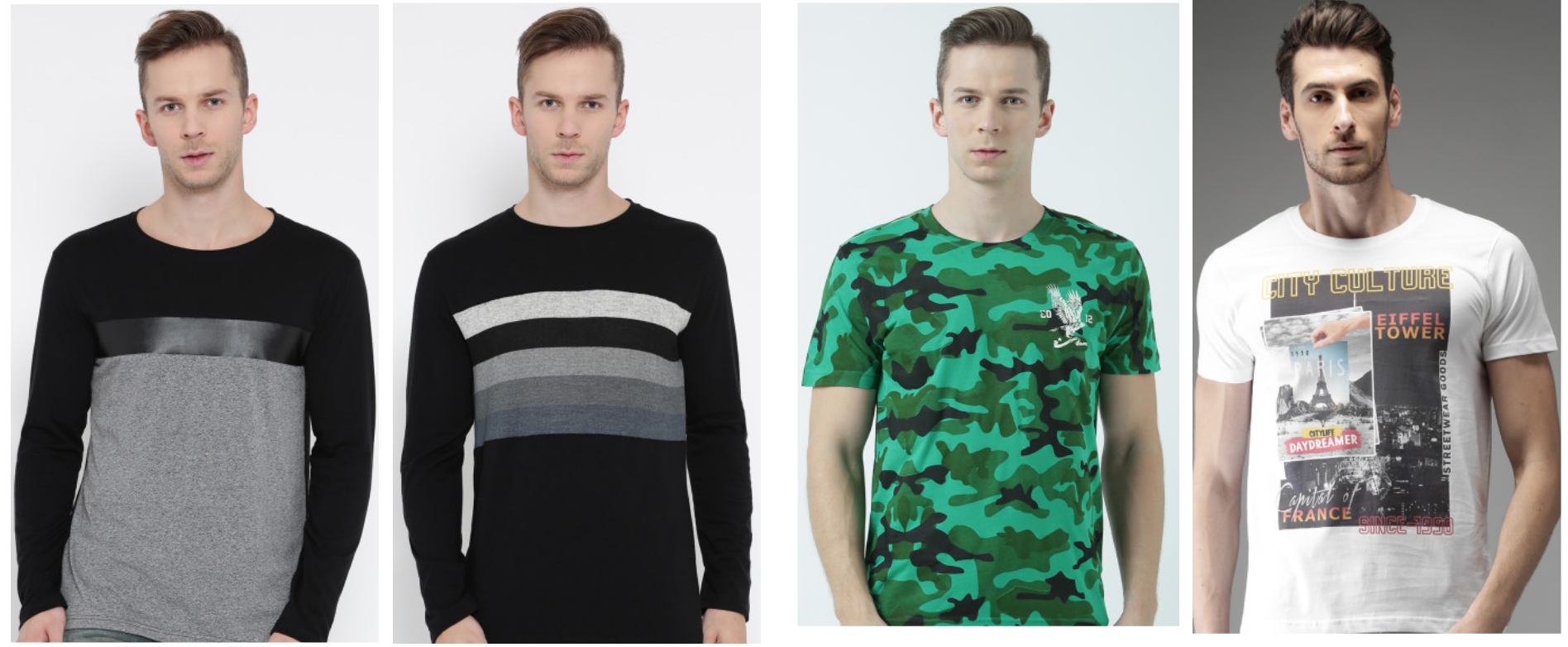}
     \caption{Ignored styles}
  \end{subfigure}
 \caption{Preference indicators from clickstream}
  \label{fig:ts}
  \end{figure}

Figure \ref{fig:ts} shows examples of preferred (row 1) and ignored items (row 2) generated using click data. These examples corroborate customer preference for polo necks over round necks. 

We also collected user session for a given time range and for women kurtas (an upper garment traditionally worn in Indian subcontinent) and found similar correlations with their independent transaction on our platform.

\section{Experiments}

From pairwise item we generated ranked list of items in the above data sets. We take top 20\% of those product and assign a positive class label and similarly bottom 20\% gets a label of negative class.  

Tables \ref{table:cl-b} shows that user-session based modelling of customer clicks which on styles which belong to similar merchandise clearly correlates with business metrics. For both classes of products (preferred and not) While the average number of impressions remains very close (approx 1\% difference), the Click-through-rate (CTR) nearly doubles in the preferred (positive) class of products. Similarly rate of sales (Quantity sold per unit of time) and revenue per unit impressions show a very substantial differences. The further support our claim on the method of user preference modelling. Next we describe few experiments which suggest that this type of modelling can be learned for inference purposes.

\begin{center}
\begin{table}[h!]
\begin{tabular}{ |c|c|c| } 
 \hline
 \bf{Metric} & \bf{Positive labelled} & \bf{Negative labelled} \\ 
 \hline
 Impressions & 410493 & 390113 \\ \hline
 CTR & 3.9\% & 2.1\%\\ \hline
 1K-revenue/Impressions & 0.412 & 0.211 \\ \hline
 rate of sale & 5.88 & 1.61 \\
 \hline
\end{tabular}
\caption{Relationship between click based feedback and business metrics}
\label{table:cl-b}
\end{table}
\end{center}

\subsection{Deep Neural Network Model based image embeddings in fashion}
For the purpose of the work presented here we have used an Image-net DNN architecture on our catalogue images (fashion catalogue at Myntra): 
Vgg16 \cite{simonyan2014very} containing about 47 million trainable parameters, reaching an ILSVRC top-5 error rate of 6.8\%. We use a pre-trained VGG16 model implemented in caffe\cite{jia2014caffe} and tap in the penultimate layer to get 4096 features vector representation of each product image after running a proprietary \cite{visenze} bounding algorithm on it. We then use PCA to get 300 dimensional feature vector representation of the product. These 300 dimensional features capture approximately 88\% of total variance. 

As mentioned in the \cite{fu2016content} features based on a pre-trained CNN (such as VGG16) are able to capture most discriminating content of a catalog images. A more complex Deep-CNNs or fine tuning on above architectures are left for future work.

\subsection{Classifiers}
We trained multiple binary (Multilayer Perceptron (MLP), RandomForestClassifier and SVM) classifiers on features described above to discriminate between articles in both buckets (preferred vs non-preferred). 

\begin{figure}[!h]
    \centering
  \begin{subfigure}[b]{0.65\linewidth}
    \includegraphics[width=\linewidth]{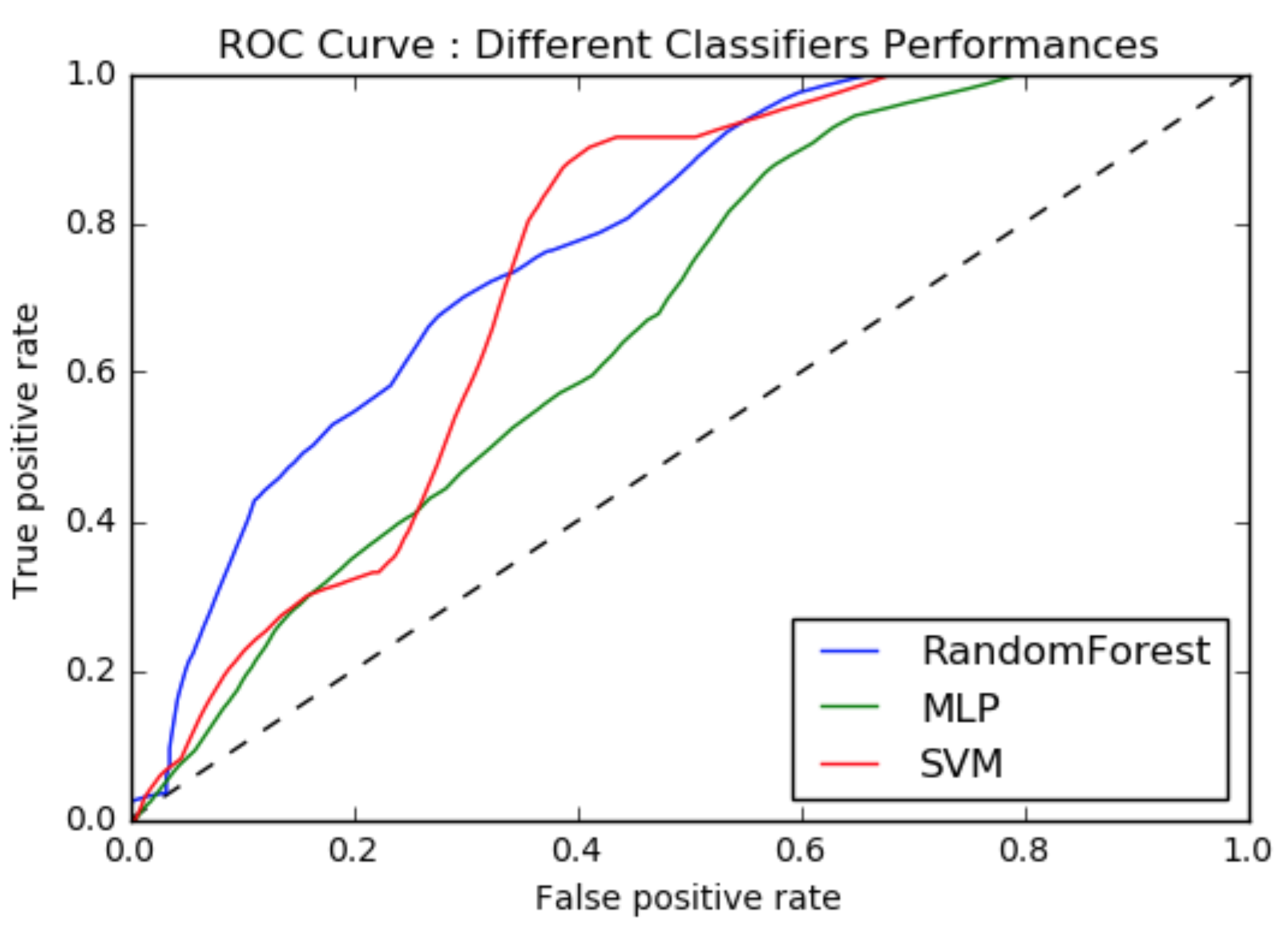}
  \end{subfigure}
 \caption{ROC curves different binary classifiers}
  \label{fig:modeling_type_comp}
  \end{figure}

Figures \ref{fig:modeling_type_comp} shows performance of various classifiers. Using these characteristic curves along with the accuracy and precision numbers we concluded that for the given feature representation it makes more sense to use ensemble model (Random Forest) as a classifier.

As described in the data preparation step we have 201666 pairs of t-shirts styles and 52082 pairs of kurtas for the classification.  We divide each of the two data sets in 75\% training and 12.5\% each in test and validation sets. Table \ref{table:data-size} shows the total exact number of data points for Training, Testing and Validation sets.

\begin{center}
\begin{table}[h!]
\begin{tabular}{ |c|c|c|c| } 
 \hline
 \bf{Article Type} & \bf{train size} & \bf{validation size} & \bf{test size}\\ 
 \hline
 T-shirts & 151250 & 25208 & 25208\\ \hline
 Kurtas & 41666 & 5208 & 5208\\ 
 \hline
\end{tabular}
\caption{Number of Pair for different article types}
\label{table:data-size}
\end{table}
\end{center}

\begin{figure}[!h]
    \centering
  \begin{subfigure}[b]{0.75\linewidth}
    \includegraphics[width=\linewidth]{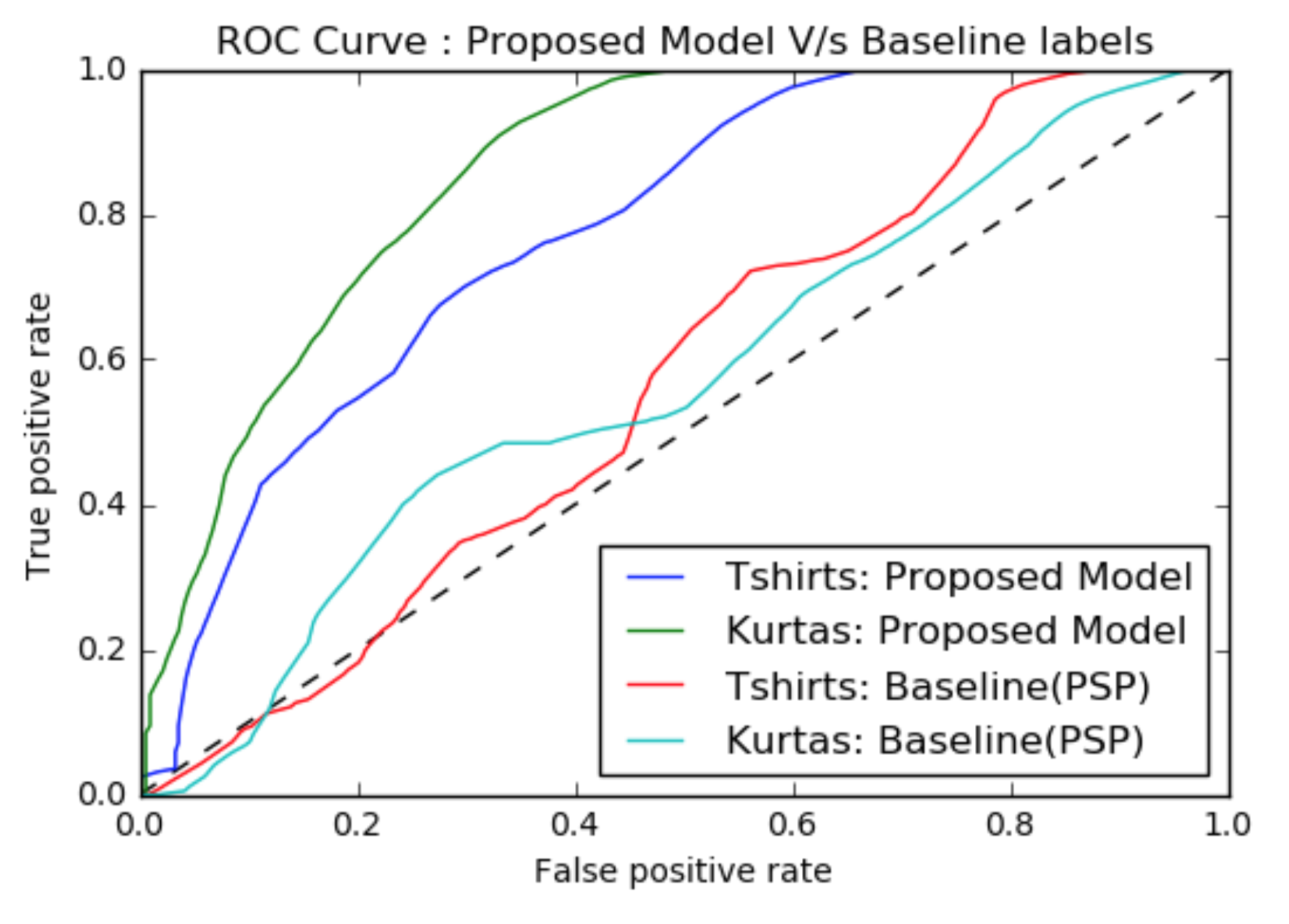}
  \end{subfigure}
 \caption{ROC curves for different item categories for proposed implicit user behaviour model vs Baseline model}
  \label{fig:category_type_comp}
  \end{figure}

Table \ref{table:clf-mod-comp} and Figures \ref{fig:category_type_comp} shows implicit user behaviour modelling as compared to the baseline PSP modelling. \cite{garg2016sales} for Men tshirts and Women Kurtas. Table \ref{table:clf-mod-comp} show significant jump in precision number from 56\% to about 64\%.

\begin{center}
\begin{table}[h!]
\begin{tabular}{ |c|c|c|c|c|c| } 
 \hline
 \bf{Article} & \bf{Model} & \bf{Acc} & \bf{AUC} & \bf{Prec} & \bf{Rec}\\ 
 \hline
 \hline
 
 \multirow{3}{4em}{T-shirts} & Implicit Feedback & 65\% & 0.66 &  \textbf{64\%} & 47\%\\ \cline{2-6}
  & Explicit feedback(\cite{garg2016sales}) & 62\% & 0.61 & 56\% & 47\%\\ \cline{2-6}
  \hline
   \multirow{3}{4em}{Kurtas} & Implicit Feedback & 66\% & 0.68 &  \textbf{63\%} & 48\%\\ \cline{2-6}
  & Explicit feedback(\cite{garg2016sales}) & 63\% & 0.64 & 56\% & 44\%\\ \cline{2-6}
  \hline
  \end{tabular}
\caption{Classifier performance for implicit vs explicit data}
\label{table:clf-mod-comp}
\end{table}
\end{center}

\subsection{Conclusion}

In this paper, We present an approach to model user sessions using
user clicks to learn user preferences for fashion product. Our approach implicitly normalizes for merchandising factors such as brand value, price, discounts, etc,  to isolate the significance of fashion content or utility . We show that implicit data correlates well with business metrics such as CTR and product sales.  We extract image features from pre-trained VGGnet to discriminate between highly clicked vs non clicked (non preferred) products. With experiments, we show that this method is able to clearly determine sellability with a significant overlap, and is therefore a good way to determine fashion preferences. \newline

\bibliographystyle{ACM-Reference-Format}
\bibliography{sample-bibliography}


\begin{thebibliography}{23}


\ifx \showCODEN    \undefined \def \showCODEN     #1{\unskip}     \fi
\ifx \showDOI      \undefined \def \showDOI       #1{#1}\fi
\ifx \showISBNx    \undefined \def \showISBNx     #1{\unskip}     \fi
\ifx \showISBNxiii \undefined \def \showISBNxiii  #1{\unskip}     \fi
\ifx \showISSN     \undefined \def \showISSN      #1{\unskip}     \fi
\ifx \showLCCN     \undefined \def \showLCCN      #1{\unskip}     \fi
\ifx \shownote     \undefined \def \shownote      #1{#1}          \fi
\ifx \showarticletitle \undefined \def \showarticletitle #1{#1}   \fi
\ifx \showURL      \undefined \def \showURL       {\relax}        \fi
\providecommand\bibfield[2]{#2}
\providecommand\bibinfo[2]{#2}
\providecommand\natexlab[1]{#1}
\providecommand\showeprint[2][]{arXiv:#2}

\bibitem[\protect\citeauthoryear{??}{vis}{[n. d.]}]%
        {visenze}
 \bibinfo{year}{[n. d.]}\natexlab{}.
\newblock In \bibinfo{booktitle}{\emph{https://www.visenze.com/}}.
\newblock


\bibitem[\protect\citeauthoryear{Datta, Joshi, Li, and Wang}{Datta
  et~al\mbox{.}}{2006}]%
        {datta2006studying}
\bibfield{author}{\bibinfo{person}{Ritendra Datta}, \bibinfo{person}{Dhiraj
  Joshi}, \bibinfo{person}{Jia Li}, {and} \bibinfo{person}{James~Z Wang}.}
  \bibinfo{year}{2006}\natexlab{}.
\newblock \showarticletitle{Studying aesthetics in photographic images using a
  computational approach}. In \bibinfo{booktitle}{\emph{European Conference on
  Computer Vision}}. Springer, \bibinfo{pages}{288--301}.
\newblock


\bibitem[\protect\citeauthoryear{Dhar, Ordonez, and Berg}{Dhar
  et~al\mbox{.}}{2011}]%
        {dhar2011high}
\bibfield{author}{\bibinfo{person}{Sagnik Dhar}, \bibinfo{person}{Vicente
  Ordonez}, {and} \bibinfo{person}{Tamara~L Berg}.}
  \bibinfo{year}{2011}\natexlab{}.
\newblock \showarticletitle{High level describable attributes for predicting
  aesthetics and interestingness}. In \bibinfo{booktitle}{\emph{Computer Vision
  and Pattern Recognition (CVPR), 2011 IEEE Conference on}}. IEEE,
  \bibinfo{pages}{1657--1664}.
\newblock


\bibitem[\protect\citeauthoryear{Fu, Li, Gao, and Wang}{Fu
  et~al\mbox{.}}{2016}]%
        {fu2016content}
\bibfield{author}{\bibinfo{person}{Ruigang Fu}, \bibinfo{person}{Biao Li},
  \bibinfo{person}{Yinghui Gao}, {and} \bibinfo{person}{Ping Wang}.}
  \bibinfo{year}{2016}\natexlab{}.
\newblock \showarticletitle{Content-based image retrieval based on CNN and
  SVM}. In \bibinfo{booktitle}{\emph{Computer and Communications (ICCC), 2016
  2nd IEEE International Conference on}}. IEEE, \bibinfo{pages}{638--642}.
\newblock


\bibitem[\protect\citeauthoryear{Garg, Banerjee, Anoop, Sreenivas, and
  Warrier}{Garg et~al\mbox{.}}{2016}]%
        {garg2016sales}
\bibfield{author}{\bibinfo{person}{Vikram Garg}, \bibinfo{person}{Rajdeep~H
  Banerjee}, \bibinfo{person}{KR Anoop}, \bibinfo{person}{T Sreenivas}, {and}
  \bibinfo{person}{Deepak Warrier}.} \bibinfo{year}{2016}\natexlab{}.
\newblock \showarticletitle{Sales Potential: Modelling Sellability of Visual
  Aesthetics of a Fashion Product}.
\newblock  (\bibinfo{year}{2016}).
\newblock


\bibitem[\protect\citeauthoryear{He, Zhang, Ren, and Sun}{He
  et~al\mbox{.}}{2016}]%
        {he2016deep}
\bibfield{author}{\bibinfo{person}{Kaiming He}, \bibinfo{person}{Xiangyu
  Zhang}, \bibinfo{person}{Shaoqing Ren}, {and} \bibinfo{person}{Jian Sun}.}
  \bibinfo{year}{2016}\natexlab{}.
\newblock \showarticletitle{Deep residual learning for image recognition}. In
  \bibinfo{booktitle}{\emph{Proceedings of the IEEE conference on computer
  vision and pattern recognition}}. \bibinfo{pages}{770--778}.
\newblock


\bibitem[\protect\citeauthoryear{He and McAuley}{He and McAuley}{2016}]%
        {he2016ups}
\bibfield{author}{\bibinfo{person}{Ruining He} {and} \bibinfo{person}{Julian
  McAuley}.} \bibinfo{year}{2016}\natexlab{}.
\newblock \showarticletitle{Ups and downs: Modeling the visual evolution of
  fashion trends with one-class collaborative filtering}. In
  \bibinfo{booktitle}{\emph{proceedings of the 25th international conference on
  world wide web}}. International World Wide Web Conferences Steering
  Committee, \bibinfo{pages}{507--517}.
\newblock


\bibitem[\protect\citeauthoryear{Isabella, Pozzani, Chen, and Gomes}{Isabella
  et~al\mbox{.}}{2012}]%
        {isabella2012influence}
\bibfield{author}{\bibinfo{person}{Giuliana Isabella},
  \bibinfo{person}{Alexandre~Ierulo Pozzani}, \bibinfo{person}{Vinicios~Anlee
  Chen}, {and} \bibinfo{person}{Murillo Buissa~Perfi Gomes}.}
  \bibinfo{year}{2012}\natexlab{}.
\newblock \showarticletitle{Influence of discount price announcements on
  consumer's behavior}.
\newblock \bibinfo{journal}{\emph{Revista de Administra{\c{c}}{\~a}o de
  Empresas}} \bibinfo{volume}{52}, \bibinfo{number}{6} (\bibinfo{year}{2012}),
  \bibinfo{pages}{657--671}.
\newblock


\bibitem[\protect\citeauthoryear{Jia, Shelhamer, Donahue, Karayev, Long,
  Girshick, Guadarrama, and Darrell}{Jia et~al\mbox{.}}{2014}]%
        {jia2014caffe}
\bibfield{author}{\bibinfo{person}{Yangqing Jia}, \bibinfo{person}{Evan
  Shelhamer}, \bibinfo{person}{Jeff Donahue}, \bibinfo{person}{Sergey Karayev},
  \bibinfo{person}{Jonathan Long}, \bibinfo{person}{Ross Girshick},
  \bibinfo{person}{Sergio Guadarrama}, {and} \bibinfo{person}{Trevor Darrell}.}
  \bibinfo{year}{2014}\natexlab{}.
\newblock \showarticletitle{Caffe: Convolutional architecture for fast feature
  embedding}. In \bibinfo{booktitle}{\emph{Proceedings of the 22nd ACM
  international conference on Multimedia}}. ACM, \bibinfo{pages}{675--678}.
\newblock


\bibitem[\protect\citeauthoryear{Ke, Tang, and Jing}{Ke et~al\mbox{.}}{2006}]%
        {ke2006design}
\bibfield{author}{\bibinfo{person}{Yan Ke}, \bibinfo{person}{Xiaoou Tang},
  {and} \bibinfo{person}{Feng Jing}.} \bibinfo{year}{2006}\natexlab{}.
\newblock \showarticletitle{The design of high-level features for photo quality
  assessment}. In \bibinfo{booktitle}{\emph{Computer Vision and Pattern
  Recognition, 2006 IEEE Computer Society Conference on}},
  Vol.~\bibinfo{volume}{1}. IEEE, \bibinfo{pages}{419--426}.
\newblock


\bibitem[\protect\citeauthoryear{Lu, Lin, Jin, Yang, and Wang}{Lu
  et~al\mbox{.}}{2014}]%
        {lu2014rapid}
\bibfield{author}{\bibinfo{person}{Xin Lu}, \bibinfo{person}{Zhe Lin},
  \bibinfo{person}{Hailin Jin}, \bibinfo{person}{Jianchao Yang}, {and}
  \bibinfo{person}{James~Z Wang}.} \bibinfo{year}{2014}\natexlab{}.
\newblock \showarticletitle{Rapid: Rating pictorial aesthetics using deep
  learning}. In \bibinfo{booktitle}{\emph{Proceedings of the 22nd ACM
  international conference on Multimedia}}. ACM, \bibinfo{pages}{457--466}.
\newblock


\bibitem[\protect\citeauthoryear{Lu, Lin, Shen, Mech, and Wang}{Lu
  et~al\mbox{.}}{2015}]%
        {lu2015deep}
\bibfield{author}{\bibinfo{person}{Xin Lu}, \bibinfo{person}{Zhe Lin},
  \bibinfo{person}{Xiaohui Shen}, \bibinfo{person}{Radomir Mech}, {and}
  \bibinfo{person}{James~Z Wang}.} \bibinfo{year}{2015}\natexlab{}.
\newblock \showarticletitle{Deep multi-patch aggregation network for image
  style, aesthetics, and quality estimation}. In
  \bibinfo{booktitle}{\emph{Proceedings of the IEEE International Conference on
  Computer Vision}}. \bibinfo{pages}{990--998}.
\newblock


\bibitem[\protect\citeauthoryear{Luo, Wang, and Tang}{Luo
  et~al\mbox{.}}{2011}]%
        {luo2011content}
\bibfield{author}{\bibinfo{person}{Wei Luo}, \bibinfo{person}{Xiaogang Wang},
  {and} \bibinfo{person}{Xiaoou Tang}.} \bibinfo{year}{2011}\natexlab{}.
\newblock \showarticletitle{Content-based photo quality assessment}. In
  \bibinfo{booktitle}{\emph{Computer Vision (ICCV), 2011 IEEE International
  Conference on}}. IEEE, \bibinfo{pages}{2206--2213}.
\newblock


\bibitem[\protect\citeauthoryear{Marchesotti, Perronnin, Larlus, and
  Csurka}{Marchesotti et~al\mbox{.}}{2011}]%
        {marchesotti2011assessing}
\bibfield{author}{\bibinfo{person}{Luca Marchesotti}, \bibinfo{person}{Florent
  Perronnin}, \bibinfo{person}{Diane Larlus}, {and} \bibinfo{person}{Gabriela
  Csurka}.} \bibinfo{year}{2011}\natexlab{}.
\newblock \showarticletitle{Assessing the aesthetic quality of photographs
  using generic image descriptors}. In \bibinfo{booktitle}{\emph{Computer
  Vision (ICCV), 2011 IEEE International Conference on}}. IEEE,
  \bibinfo{pages}{1784--1791}.
\newblock


\bibitem[\protect\citeauthoryear{Marchesotti, Perronnin, and
  Meylan}{Marchesotti et~al\mbox{.}}{2013}]%
        {marchesotti2013learning}
\bibfield{author}{\bibinfo{person}{Luca Marchesotti}, \bibinfo{person}{Florent
  Perronnin}, {and} \bibinfo{person}{France Meylan}.}
  \bibinfo{year}{2013}\natexlab{}.
\newblock \showarticletitle{Learning beautiful (and ugly) attributes.}. In
  \bibinfo{booktitle}{\emph{BMVC}}, Vol.~\bibinfo{volume}{7}.
  \bibinfo{pages}{1--11}.
\newblock


\bibitem[\protect\citeauthoryear{M{\"u}ller, Clough, Deselaers, Caputo, and
  CLEF}{M{\"u}ller et~al\mbox{.}}{2010}]%
        {muller2010experimental}
\bibfield{author}{\bibinfo{person}{Henning M{\"u}ller}, \bibinfo{person}{Paul
  Clough}, \bibinfo{person}{Thomas Deselaers}, \bibinfo{person}{Barbara
  Caputo}, {and} \bibinfo{person}{Image CLEF}.}
  \bibinfo{year}{2010}\natexlab{}.
\newblock \showarticletitle{Experimental evaluation in visual information
  retrieval}.
\newblock \bibinfo{journal}{\emph{The Information Retrieval Series}}
  \bibinfo{volume}{32} (\bibinfo{year}{2010}), \bibinfo{pages}{1--554}.
\newblock


\bibitem[\protect\citeauthoryear{Murray, Marchesotti, and Perronnin}{Murray
  et~al\mbox{.}}{2012}]%
        {murray2012ava}
\bibfield{author}{\bibinfo{person}{Naila Murray}, \bibinfo{person}{Luca
  Marchesotti}, {and} \bibinfo{person}{Florent Perronnin}.}
  \bibinfo{year}{2012}\natexlab{}.
\newblock \showarticletitle{AVA: A large-scale database for aesthetic visual
  analysis}. In \bibinfo{booktitle}{\emph{Computer Vision and Pattern
  Recognition (CVPR), 2012 IEEE Conference on}}. IEEE,
  \bibinfo{pages}{2408--2415}.
\newblock


\bibitem[\protect\citeauthoryear{Nishiyama, Okabe, Sato, and Sato}{Nishiyama
  et~al\mbox{.}}{2011}]%
        {nishiyama2011aesthetic}
\bibfield{author}{\bibinfo{person}{Masashi Nishiyama},
  \bibinfo{person}{Takahiro Okabe}, \bibinfo{person}{Imari Sato}, {and}
  \bibinfo{person}{Yoichi Sato}.} \bibinfo{year}{2011}\natexlab{}.
\newblock \showarticletitle{Aesthetic quality classification of photographs
  based on color harmony}. In \bibinfo{booktitle}{\emph{Computer Vision and
  Pattern Recognition (CVPR), 2011 IEEE Conference on}}. IEEE,
  \bibinfo{pages}{33--40}.
\newblock


\bibitem[\protect\citeauthoryear{O'Donovan, Agarwala, and Hertzmann}{O'Donovan
  et~al\mbox{.}}{2011}]%
        {o2011color}
\bibfield{author}{\bibinfo{person}{Peter O'Donovan}, \bibinfo{person}{Aseem
  Agarwala}, {and} \bibinfo{person}{Aaron Hertzmann}.}
  \bibinfo{year}{2011}\natexlab{}.
\newblock \showarticletitle{Color compatibility from large datasets}. In
  \bibinfo{booktitle}{\emph{ACM Transactions on Graphics (TOG)}},
  Vol.~\bibinfo{volume}{30}. ACM, \bibinfo{pages}{63}.
\newblock


\bibitem[\protect\citeauthoryear{Russakovsky, Deng, Su, Krause, Satheesh, Ma,
  Huang, Karpathy, Khosla, Bernstein, et~al\mbox{.}}{Russakovsky
  et~al\mbox{.}}{2015}]%
        {russakovsky2015imagenet}
\bibfield{author}{\bibinfo{person}{Olga Russakovsky}, \bibinfo{person}{Jia
  Deng}, \bibinfo{person}{Hao Su}, \bibinfo{person}{Jonathan Krause},
  \bibinfo{person}{Sanjeev Satheesh}, \bibinfo{person}{Sean Ma},
  \bibinfo{person}{Zhiheng Huang}, \bibinfo{person}{Andrej Karpathy},
  \bibinfo{person}{Aditya Khosla}, \bibinfo{person}{Michael Bernstein},
  {et~al\mbox{.}}} \bibinfo{year}{2015}\natexlab{}.
\newblock \showarticletitle{Imagenet large scale visual recognition challenge}.
\newblock \bibinfo{journal}{\emph{International Journal of Computer Vision}}
  \bibinfo{volume}{115}, \bibinfo{number}{3} (\bibinfo{year}{2015}),
  \bibinfo{pages}{211--252}.
\newblock


\bibitem[\protect\citeauthoryear{Simonyan and Zisserman}{Simonyan and
  Zisserman}{2014}]%
        {simonyan2014very}
\bibfield{author}{\bibinfo{person}{Karen Simonyan} {and}
  \bibinfo{person}{Andrew Zisserman}.} \bibinfo{year}{2014}\natexlab{}.
\newblock \showarticletitle{Very deep convolutional networks for large-scale
  image recognition}.
\newblock \bibinfo{journal}{\emph{arXiv preprint arXiv:1409.1556}}
  (\bibinfo{year}{2014}).
\newblock


\bibitem[\protect\citeauthoryear{Szegedy, Liu, Jia, Sermanet, Reed, Anguelov,
  Erhan, Vanhoucke, Rabinovich, et~al\mbox{.}}{Szegedy et~al\mbox{.}}{2015}]%
        {szegedy2015going}
\bibfield{author}{\bibinfo{person}{Christian Szegedy}, \bibinfo{person}{Wei
  Liu}, \bibinfo{person}{Yangqing Jia}, \bibinfo{person}{Pierre Sermanet},
  \bibinfo{person}{Scott Reed}, \bibinfo{person}{Dragomir Anguelov},
  \bibinfo{person}{Dumitru Erhan}, \bibinfo{person}{Vincent Vanhoucke},
  \bibinfo{person}{Andrew Rabinovich}, {et~al\mbox{.}}}
  \bibinfo{year}{2015}\natexlab{}.
\newblock \showarticletitle{Going deeper with convolutions}. Cvpr.
\newblock


\bibitem[\protect\citeauthoryear{Zhang, Zhang, Zhang, Lai, Liu, Zhang, and
  Ma}{Zhang et~al\mbox{.}}{2015}]%
        {zhang2015daily}
\bibfield{author}{\bibinfo{person}{Yongfeng Zhang}, \bibinfo{person}{Min
  Zhang}, \bibinfo{person}{Yi Zhang}, \bibinfo{person}{Guokun Lai},
  \bibinfo{person}{Yiqun Liu}, \bibinfo{person}{Honghui Zhang}, {and}
  \bibinfo{person}{Shaoping Ma}.} \bibinfo{year}{2015}\natexlab{}.
\newblock \showarticletitle{Daily-aware personalized recommendation based on
  feature-level time series analysis}. In \bibinfo{booktitle}{\emph{Proceedings
  of the 24th international conference on world wide web}}. International World
  Wide Web Conferences Steering Committee, \bibinfo{pages}{1373--1383}.
\newblock


\end{thebibliography}

\end{document}